\documentclass[preprint,12pt]{elsarticle}

\usepackage{amssymb}
\usepackage{amsmath}
\usepackage{psfrag}
\usepackage{epsfig}
\usepackage{graphicx,subfigure}
\usepackage{color}
\usepackage{algorithm}
\usepackage{algorithmic}
\usepackage{rotating}

\journal{Neurocomputing}
\begin{document}

\begin{frontmatter}

\title{Convergence analysis of beetle antennae search algorithm and its applications \tnoteref{SNSF}}
\tnotetext[SNSF]{This work is supported by the Hong Kong Research
Grants Council Early Career Scheme (with number 25214015), and by
Departmental General Research Fund of Hong Kong Polytechnic
University (with number G.61.37.UA7L), by the National Natural
Science Foundation of China (with numbers 61622308 and 61873206), by
the Fok Ying-Tong Education Foundation (with number 161058), and
also by the Science and Technology on Space Intelligent Control
Laboratory (with number ZDSYS-2017-05)}
\author[myfirstaddress]{Yinyan Zhang}
%\ead{yinyan.zhang@connect.polyu.hk}
\author[myfirstaddress]{Shuai Li\corref{mycorrespondingauthor}}
\author[mysecondaddress]{Bin Xu}
\cortext[mycorrespondingauthor]{Corresponding author.}
\ead{shuaili@polyu.edu.hk.}
\address[myfirstaddress]{Department of
Computing, The Hong Kong Polytechnic University, Hong Kong, China}
\address[mysecondaddress]{School of Automation, Northwestern Polytechnical
University, Xi¡¯an, China, 710072}

\begin{abstract}
The  beetle antennae search algorithm was recently proposed and
investigated for solving global optimization problems. Although the
performance of the algorithm and its variants were shown to be
better than some existing meta-heuristic algorithms, there is still
a lack of convergence analysis. In this paper, we provide
theoretical analysis on the convergence of the beetle antennae
search algorithm. We test the performance of the BAS algorithm via
some representative benchmark functions. Meanwhile, some
applications of the BAS algorithm are also presented.
\end{abstract}


\end{frontmatter}

\section{Introduction}
As a meta-heuristic algorithm, the beetle antennae search (BAS)
algorithm was proposed by Jiang and Li \cite{basba}. The design of
the algorithm was inspired by the behaviors of beetles when seeking
for a mate. The performance of the BAS algorithm has been evaluated
in various applications. Zhu {\it et al.} \cite{anbas} applied BAS
algorithm to multiobjective energy management in microgrids which
adopts minimum operation cost and minimum pollutant treatment cost
as its objectives under the constraints of time-of-use price and
energy storage status. Yin and Ma \cite{asfcm} proposed  an
aggregation service chain mapping plan based on an improved BAS
algorithm for network resources allocation, which consumes less
computing resources and has excellent performance in key mapping
costs and network latency. Wang {\it et al.} \cite{rosee} applied
the BAS algorithm to improve the accuracy of spatial straightness
assessment, showing a faster convergence and better accuracy. Sun
{\it et al.} \cite{onnub} used the BAS algorithm to train a neural
network, which was further applied to the prediction of the
unconfined compressive strength of jet grouting coalcretes, which
showed a better performance than multiple regression, logistic
regression, and support vector machine. Lin {\it et al.}
\cite{darod} utilized the BAS algorithm to the tuning of a PID
controller for DC motors, which led to a smaller overshooting and a
faster responding speed when the load and disturbance changes
compared with a traditional PID controller. Sun {\it et al.}
\cite{doymo} used the BAS algorithm to tune the hyperparameters of
support vector machine for the determination of Young's modulus of
jet grouted coalcretes. Compared with other algorithms, the method
proposed by Sun {\it et al.} is less time-consuming and more
accurate with a lower cost. Sun {\it et al.} \cite{popau} adopted
the BAS algorithm to tune a support vector regression model for the
prediction of permeability and unconfined compressive strength of
pervious concretes, leading to a high prediction accuracy. The above
works showed that the convergence of the BAS algorithm is fast, the
implementation of the BAS algorithm is simple, and the probability
of the BAS algorithm to be trapped in local optimum is small.
Recently, the combinations of BAS with particle swarm optimization
(PSO) were also reported. Chen {\it et al.} \cite{bsofs} proposed a
beetle swarm optimization (BSO) algorithm by combining the beetle
antennae search (BAS) algorithm with the standard PSO algorithm,
where the update rule of each particle follows BAS. The algorithm
was also adopted to solve the wireless sensor network coverage
problem, showing a better performance than the standard PSO
\cite{aopso}. The BSO algorithm was then adopted to solve an
investment portfolio problem. The combination of BAS with BSO was
also proposed in \cite{ahomo}, which has a better performance than
standard BSO.

While the BAS algorithm has been found to be efficient and effective
in solving many optimization problems, there is still a lack of
theoretical guarantee. Motivated by this fact, in this paper, we aim
at providing convergence analysis on the BAS algorithm. We will also
validate the performance of the algorithm with some typical
examples. The contributions of this paper are listed as follows:
\begin{itemize}
\item[1)] The theoretical guarantee for the performance of the BAS
algorithm is provided.
\item[2)] The quantitive analysis on the performance of the BAS
algorithm for seven representative test functions are conducted
based on the successful rate measure.
\item[3)] The performance of the BAS algorithm in engineering
applications is tested.
\end{itemize}
The rest of this paper is organized as follows. In Section
\ref{sec.2}, we revisit the BAS algorithm, followed by the
theoretical analysis on Section \ref{sec.3}. Then, we test the
performance of the BAS algorithm through numerical experiments in
Section \ref{sec.4}. The performance of the BAS algorithm is also
tested by three engineering problems in Section \ref{sec.appl}.
Conclusions are given in Section \ref{sec.5}.

\section{Algorithm description}\label{sec.2}
In this section, we review the BAS algorithm.

Consider the minimization problem of function
$f(\mathbf{x})\in\mathbb{R}$ with the decision variable being
$\mathbf{x}=[x_1,x_2,\cdots,x_n]^\text{T}$.

{\it Assumption 1:} The optimal solution to the minimization problem
of $f(\mathbf{x})$ exists.

The BAS algorithm treats the  decision variable as the location of
the centroid position of a beetle in the $n$-dimensional space. To
minimize the function $f$, the behavior of the beetle is described
as follows according to the BAS algorithm \cite{basba}:
\begin{equation}\label{eq.1}
\mathbf{x}^{k+1}=\mathbf{x}^k-\delta^k\mathbf{b}~\text{sgn}(f(\mathbf{x}^k_l)-f(\mathbf{x}^k_r)),
\end{equation}
where $\mathbf{x}^k_l$ and $\mathbf{x}^k_r$ denote the location of
the left tentacle and the right tentacle of the beetle at time
instant $k$, respectively; $\delta^k$ denotes the step size of
searching; $\mathbf{b}$ denotes a direction vector, which is random,
and set as follows:
\begin{equation}\label{eq.b}
\mathbf{b}=\frac{\text{rnd}(n,1)}{\|\text{rnd}(n,1)\|_2},
\end{equation}
with $\|\cdot\|$ denoting the two-norm operator and
$\text{rnd}(n,1)$ denotes a randomly generated $n$-dimensional
vector; $\text{sgn}(\cdot)$ is the sign function. The locations of
left and right tentacles are given as follows:
\begin{equation}\label{eq.lr}
\begin{aligned}
\mathbf{x}_l&=\mathbf{x}^k+d^k\mathbf{b},\\
\mathbf{x}_r&=\mathbf{x}^k-d^k\mathbf{b}.
\end{aligned}
\end{equation}
In addition, in the BAS algorithm, it is suggested to set
\begin{equation}\label{eq.detald}
\begin{aligned}
\delta^k&=\alpha \delta^{k-1}+0.001,\\
 d^\text{k}&=c* d^\text{k-1}+d_0
\end{aligned}
\end{equation}
with $c>0\in\mathbb{R}$, $\alpha\in(0,1)$,
$\delta^0>0\in\mathbb{R}$, and $d_0>0\in\mathbb{R}$.

If the searching ranging is defined in a closed set
$\Omega\in\mathbb{R}^n$, then the BAS algorithm is modified as
\cite{basba}:
\begin{equation}\label{eq.2}
\mathbf{x}^{k+1}=P_\Omega(\mathbf{x}^k-\delta^k\mathbf{b}~\text{sgn}(f(\mathbf{x}^k_l)-f(\mathbf{x}^k_r))),
\end{equation}
where $P_\Omega(\cdot)$ denotes the projection operator. Evidently,
(\ref{eq.1}) is a special case of (\ref{eq.2}) by setting
$\Omega=R^n$.

The basic BAS algorithm is given in Algorithm \ref{algorithm.a}.

\begin{algorithm}
    \caption{BAS algorithm for global minimization}
    \label{algorithm.a}
    \begin{algorithmic}
    \REQUIRE Objective function $f(\mathbf{x})$, and values of
    parameters $\alpha$, $c$, $\delta^0$, $d_0$, $\mathbf{x}^0$, and searching
    set $\Omega$
    \ENSURE Optimal solution $\mathbf{x}_{bst}$ and optimal function value $f_{bst}$.
    \STATE Initialize $f_{bst}$ to be $f(\mathbf{x}^0)$
    \STATE Initialize $\mathbf{x}_{bst}$ to be $\mathbf{x}^0$
    \WHILE{($k<K_{max}$) or (stop criterion)}
    \STATE Generate $\mathbf{b}$ according to (\ref{eq.b})
    \STATE Calculate $\mathbf{x}^k_l$ and $\mathbf{x}^k_r$ according
    to (\ref{eq.lr})
    \STATE Calculate $\mathbf{x}^{k+1}$ according to (\ref{eq.2})
    \IF{$f(\mathbf{x}^{k+1}) < f_{bst}$}
    \STATE $f_{bst}=f(\mathbf{x}^{k+1}), \mathbf{x}_{bst}=\mathbf{x}^{k+1}$
    \ENDIF
    \ENDWHILE
    \end{algorithmic}
\end{algorithm}

\section{Convergence analysis}\label{sec.3}
In this section, convergence analysis for the BAS algorithm is
provided. We first give the definition of convergence as follows.

{\it Definition 1} \cite{mbrst}: (Convergence with probability 1)
Convergence with probability 1 means that with probability 1 a
monotone sequence $\{f(\mathbf{x})\}^\infty_{k=1}$ which converges
to the infimum of $f$ is obtained on $\Omega$.

The convergence analysis is based on Definition 1. Before moving to
the analysis, for the sake of illustration, let
$\mathbf{x}^k_{bst}=\min_{\mathbf{x}^j}\{f(\mathbf{x}^j)\}$ with
$j=0,1,\cdots,k$ and $f^k_{bst}=f(\mathbf{x}^k_{bst})$.

{\it Lemma 1:} For the BAS algorithm, $f^k_{bst}$ is not increasing.

{\it Proof:} According to Algorithm \ref{algorithm.a}, at each
instant $k$, if $f(\mathbf{x}^{k+1}) < f_{bst}$, then
$f_{bst}=f(\mathbf{x}^{k+1})$. Note that the initial value of
$f_{bst}$ to be extremely large. As a result, the BAS algorithm
guarantees that $f^k_{bst}$ is not increasing.

Lemma gives a determined conclusion that the BAS algorithm will not
diverge in the long term.

{\it Theorem 1:} Given that the parameters are properly set, the BAS
algorithm is convergent with probability 1.

{\it Proof:} Suppose that the parameters of the BAS algorirthm are
properly set such that at each time instant $k$, the probability of
$P_\Omega(\mathbf{x}^k+\delta^k\mathbf{b}~\text{sgn}(f(\mathbf{x}^k_r)-f(\mathbf{x}^k_l)))$
located on the optimal solution $\mathbf{x}^*$ to the minimization
problem of $f$ is larger than 0. Let $p_k$ denotes the probability
that at time instant $k$, $\mathbf{x}^k$ is not located on
$\mathbf{x}^*$. Then, we have
\begin{equation*}
p(\mathbf{x}^k_{bst}=\mathbf{x}^*)>=1-p_0p_1\cdots p_k.
\end{equation*}
Note that $0\leq p_k<1$ by the above assumption. Thus,
\begin{equation*}
\lim_{k\rightarrow+\infty} (1-p_0p_1 \cdots  p_k )= 1-
\lim_{k\rightarrow+\infty} p_0p_1 \cdots  p_k = 1.
\end{equation*}
Note that $$p(\mathbf{x}^k_{bst}=\mathbf{x}^*)\leq1.$$ Thus, by the
squeeze theorem, we further have $$\lim_{t\rightarrow+\infty}
p(\mathbf{x}^k_{bst}=\mathbf{x}^*) =1.$$ The proof is complete.
$\hfill\Box$

Theorem 1 shows that by properly choosing the step size, we can
guarantee that the BAS algorithm is asymptotic convergent will
probability 1. This conclusion is important. Firstly, it shows that
the BAS algorithm can converge under a condition about its step
size. Secondly, in practice, this theorem also helps us identify the
problem about why the BAS algorithm may not have a good solution
performance when facing certain functions, which is a general issue
in most bio-inspired algorithms.

\begin{sidewaystable}
%\begin{table*}[t]
  \centering
  \caption{List of test functions}
  \begin{tabular} {lc  cc   c}
  \hline
   function                                      &dimension            & global minima & \\
\hline
    $f_1(\mathbf{x})=\displaystyle\sqrt{\sum^n_{i=1}x^2_i}$                               &$n=30$                 &$f^*_1=0$ at $\mathbf{x}^*=\mathbf{0}\in\mathbb{R}^{n}$                        \\
    $f_2(\mathbf{x})=\displaystyle\sum^n_{i=1}|x_i|+\prod^n_{i=1}|x_i|$            &$n=20$          &$f^*_2=0$ at $\mathbf{x}^*=\mathbf{0}\in\mathbb{R}^{n}$ \\
    $f_3(\mathbf{x})=\displaystyle\sum^{n-1}_{i=1}(100(x_{i+1}-x^2_i)^2+(x_i-1)^2)$ &$n=10$     &$f^*_3=0$ at $\mathbf{x}^*=\mathbf{1}\in\mathbb{R}^{n}$ \\
    $f_4(\mathbf{x})=\displaystyle-20\exp(-0.2\sqrt{\frac{1}{n}\sum^n_{i=1}x^2_i})-\exp(\frac{1}{n}\sum^n_{i=1}\cos(2\pi
    x_i))+20+\exp(1)$ &$n=10$      &$f^*_4=0$ at
    $\mathbf{x}^*=\mathbf{0}\in\mathbb{R}^{n}$\\
    $f_5(\mathbf{x})=\displaystyle1+\frac{1}{4000}\sum^n_{i=1}x^2_i-\prod^n_{i=1}\frac{\cos
    x_i}{\sqrt{i}}$ & $n=10$ &$f^*_5=0$ at $\mathbf{x}^*=\mathbf{0}\in\mathbb{R}^{n}$   \\
    $f_6(\mathbf{x})=\displaystyle(\sum^n_{i=1}|x_i|)\exp(-\sum^n_{i=1}\sin
    x^2_i)$ & $n=5$ & $f^*_6=0$ at
    $\mathbf{x}^*=\mathbf{0}\in\mathbb{R}^{n}$\\
    $f_7(\mathbf{x})=\displaystyle\sum^n_{i=1}x^2_i+(0.5\sum^n_{i=1}ix_i)^2+(0.5\sum^n_{i=1}ix_i)^4$
    & $n=20$ & $f^*_1=0$ at
    $\mathbf{x}^*=\mathbf{0}\in\mathbb{R}^{n}$\\
    \hline
\end{tabular} \label{tab1}
%\end{table*}
\end{sidewaystable}

\section{Illustrative examples}\label{sec.4}

In this section, we provide some illustrative examples to show the
performance of the BAS algorithm.

There are many criteria   for evaluating the performance of
bio-inspired algorithms for solving optimization problem, such as
the success rate and number of function evaluations. In this paper,
we adopt the success rate to evaluate the performance of the BAS
algorithm, which is defined as follows \cite{def}:
\begin{equation}\label{eq.suss}
\text{success rate} = \frac{N_\text{success}}{N_{all}},
\end{equation}
where $N_\text{success}$ denotes the number of successful trials and
$N_\text{all}$ denotes the total number of trials. A trial is
considered to be successful if the following inequality is
satisfied:
\begin{equation}
\sum^n_{i=1}(x_{bsti}-x^*_i)^2\leq(UB-LB)\times10^{-4},
\end{equation}
where $UB$ denotes the identical upper bound and $LB$ denotes the
identical lower bound of the elements in $\mathbf{x}$.

Seven test functions adopted from \cite{def,yang1,yang2} are
considered in this paper. The function expressions, dimensions, and
the corresponding global optima are listed on Table \ref{tab1}. The
variable bounds for the optimum searching of each function are -10
to 10 (i.e., $LB=-10$ and $UB=10$) for each variable for all the
functions, except that, for function $f_6$, we have $LB=-2\pi$ and
$UB=2\pi$. ALL the test functions have a unique global optimum in
the given search regions such that we can easily use (\ref{eq.suss})
to evaluate the performance of the BAS algorithm. These test
functions are selected due to their representative properties. For
example, $f_4$ called Griewank's function is highly multimodal,
meaning that it has many local minima. For each function, the
maximum number of iterations in each run is set to $10^5$ (i.e.,
$K_{max}$ is set to $10^5$) and each function is tested for 100 runs
by using the BAS algorithm. The initial value of each element in
$\mathbf{x}$ for each
 test function is randomly generated with a uniformly distribution.
 The initial values of $\delta$ is set to 10 and the initial value
 of $d$ is set to $UB$ for all test functions.

 The test results and parameter settings are shown in Table
 \ref{tab2}. As seen from Table \ref{tab2}, the successful rate of
 the BAS algorithm is relatively high for the test functions. For
 example, for functions $f_1$, $f_5$, $f_6$, and $f_7$, the
 successful rate is 100. The lowest successful rate of the BAS
 algorithm is 80, which is for function $f_4$. This is due to the
 aforemensioned fact, i.e., $f_4$ is highly multimodal. It is worth pointing out that
 the successful rate depends on the parameter setting. However,
 currently, the parameters are set manually. Thus, better results
 could be obtained if some automatic paramter tuning methods are
 used. As seen from Table \ref{tab2}, the standard deviation of the
 obtained optimal function value is relatively row except for
 functions $f_2$ and $f_3$. The reason for this could be that there
 are some sharp regions in the two functions. Reagarding the
 best function optima obtained by the BAS algorithm, we can see that
 the differences between the obtained ones and the theoretical ones
 are about $10^{-2}$ for most functions. This is related to the
 setting of step size. Normally, if we want to have a more accurate
 optimum, we need to have a smaller step size, which generally will
 lead to larger consumption of computational resouces. In other
 words, there is a trade-off between accuracy and efficiency. Here,
 our evalution criterion is the successful rate, which serves as a
 trade-off criterion. To sum up, the BAS algorithm has a good
 performance for finding global optima of functions, regardless of
 whether they are multimodal or not.

\begin{sidewaystable}
%\begin{table*}[t]
  \centering
  \caption{Test results and paramter settings of BAS algorithm for the test functions shown in Table \ref{tab1}}
  \begin{tabular} {lc  cc   cc}
  \hline
   function                                      &parameter setting           & successful rate & best $f_{bst}$ & average $f_{bst}$ & standard deviation of $f_{bst}$\\
\hline
    $f_1(\mathbf{x})$                               &$ \alpha =0.94, d_0=0.001,  c=0.94$
    & $100$ & $0.0271$ & $ 0.0311$ & $ 0.0016$
    \\
      $f_2(\mathbf{x})$                               &$ \alpha =0.95, d_0=0.001,  c=0.94$
    & $96$ & $0.0708$ & $ 2.8432$ & $ 15.0748$
    \\
          $f_3(\mathbf{x})$                               &$ \alpha =0.7, d_0=0.001,  c=0.7$
    & $82$ & $5.3561e-04$ & $0.5659$ & $  1.4076$
    \\
              $f_4(\mathbf{x})$                               &$ \alpha =0.97, d_0=0.01,  c=0.97$
    & $80$ & $ 0.0122$ & $0.4377$ & $  0.8607$
    \\
                  $f_5(\mathbf{x})$                               &$ \alpha =0.94, d_0=0.001,  c=0.94$
    & $100$ & $  0.9995$ & $0.9995$ &$  8.0206e-09$
    \\
                      $f_6(\mathbf{x})$                               &$ \alpha =0.96, d_0=0.1,  c=0.96$
    & $100$ & $   0.0032$ & $ 0.0085$ &$    0.0019$
    \\
                          $f_7(\mathbf{x})$                               &$ \alpha =0.8, d_0=0.01,  c=0.8$
    & $100$ & $    3.3053e-04$ & $ 6.3522e-04$ &$    1.1920e-04$
    \\
    \hline
\end{tabular} \label{tab2}
%\end{table*}
\end{sidewaystable}

\section{Applications}\label{sec.appl}
In this section, we show the application of the BAS algorithm to
some engineering problems.

\begin{sidewaystable}
%\begin{table*}[t]
  \centering
  \caption{Best result of the BAS algorithm among the 1000 runs for solving the spring design problem and the best result obtained by the Bat algorithm in \cite{appl1}}
  \begin{tabular} {lcc  cc   cccc}
  \hline
   algorithm  &  $f(\mathbf{x})$                                &$W$           & $D$ & $L$ & $g_1(\mathbf{x})$ & $g_2(\mathbf{x})$ &$g_3(\mathbf{x})$ & $g_4(\mathbf{x})$\\
\hline
    BAS algorithm & $ 0.010894$                               & 0.050000
    &  0.360419 & $ 10.090624$ & $ -0.052996$ & $ -4.357457$ & -0.726387 &-0.035687\\
    Bat algorithm& 0.012665 & 0.051690 & 0.356750 & 11.287126 & / & / & / & / \\
    \hline
\end{tabular} \label{tab3}
%\end{table*}
\end{sidewaystable}

\subsection{Spring design problem}
The optimal design problem of a tensional and compressional spring
is described as follows \cite{appl1,appl2}:
\begin{equation*}
\begin{aligned}
\text{min}&~f(\mathbf{x})=(L+2)W^2D, \\
\text{subject to~~}& g_1(\mathbf{x})=1-\frac{D^3L}{71785W^4}\leq0,\\
&g_2(\mathbf{x})=1-\frac{140.45W}{D^2L}\leq0,\\
&g_3(\mathbf{x})=\frac{2(W+D)}{3}-1\leq0,\\
&g_4(\mathbf{x})=\frac{D(4D-W)}{W^3(12566D-W)}+\frac{1}{5108W^2}-1\leq0,\\
&0.05\leq W \leq2.0,\\
&0.25\leq D\leq1.3,\\
&2.0\leq L \leq 15.0,\\
\end{aligned}
\end{equation*}
where $f(\mathbf{x})$ is the weight of the spring which needs to be
minimized, $W$ denotes the wire diameter, $D$ denotes the mean coil
diameter, and $L$ denotes the length or the number of coils. The
contraints are related to the maximum shear stress, minimum
deflection, etc. The details can be found in \cite{appl2}. We first
convert the problem to a form that can be addressed by the BAS
algroithm by using the penalty method:
\begin{equation*}
\begin{aligned}
\text{min}&~f(\mathbf{x})=(L+2)W^2D+\rho h_i(\mathbf{x}),\\
\text{subject to~~}&0.05\leq W \leq2.0,\\
&0.25\leq D\leq1.3,\\
&2.0\leq L \leq 15.0,\\
\end{aligned}
\end{equation*}
where
\begin{equation}
h_i(\mathbf{x})=\text{max}(0,g_i(\mathbf{x})),
\end{equation}
and $\rho$ is called the penalty paramter. In the numerical
experiment, we set $\rho=10^5$, and the parameters of the BAS
algorithm is set as   $\alpha = 0.8$, $d_0=0.01$, and $c=0.8$. We
run the BAS algorithm for 1000 times and in each run the initial
values of $\mathbf{x}$ are set by following the rules in the
previous section with $K_{max}=1000$. The best result among the 1000
runs are shown in Table \ref{tab3}, where all the constraints are
satisfied. Obviously, the result is better than the best result
abtained by the Bat algorithm discussed in \cite{appl1}, for which
the optimum is $0.012665$.

\begin{sidewaystable}
%\begin{table*}[t]
  \centering
  \caption{Best result of the BAS algorithm among the 100 runs for solving the speed reducer problem and existing results}
  \begin{tabular} {llll}
  \hline
     &                           BAS algorithm             &bat algorithm \cite{appl1}      &deterministic technique \cite{apl3}\\
\hline
$B$                       &  3.501597128660806          &  3.5                              &3.5\\
$H$                       &  0.7                        &  0.7                              &0.7\\
$Z$                       &  17                         &  17                               &17\\
$L_1$                     &  8.104555092323999          &  7.3                              &7.3\\
$L_2$                     &  8.021701619497760          &  7.8                              &7.7153190\\
$D_1$                     &  3.353618456239036          &  3.34336445                       &3.350282\\
$D_2$                     &  5.291060245756827          &  5.285350625                      &5.286654\\
$g_1(\mathbf{x})$         &  -0.074337680917883         &  -0.073915280397873               &-0.073915280397873\\
$g_2(\mathbf{x})$         &  -0.198364331513853         &  -0.197998527141949               &-0.197998527141949\\
$g_3(\mathbf{x})$         &  -0.317436155693268         &  -0.495055034120807               &-0.499212509394955\\
$g_4(\mathbf{x})$         &  -0.893183330622976         &  -0.901372291570736               &-0.904643904804176\\
$g_5(\mathbf{x})$         &  -0.001627492443412         &  $0.006159299578992\star$         &-6.029273083829612e-05\\
$g_6(\mathbf{x})$         &  -0.002436220272286         &  $7.565860087876963e-04\star$     &2.636860652049933e-07$\star$\\
$g_7(\mathbf{x})$         &  -0.702500000000000         &  ${-0.702500000000000}$           &-0.702500000000000\\
$g_8(\mathbf{x})$         &  -4.561143392921574e-04     &  0                                &0\\
$g_9(\mathbf{x})$         &  -0.583143198968952         &   -0.583333333333333              &-0.583333333333333\\
$g_{10}(\mathbf{x})$      &  -0.144872530890374         &  -0.052733332191781               &-0.051311917808219\\
$g_{11}(\mathbf{x})$      &  -0.037589948301284         &  -0.011040296474359               &5.184490747822679e-08$\star$\\
$f(\mathbf{x})$           & 3.012610927770214e+03       &  2.993758748042880e+03            &2.994487910428388e+03\\
    \hline
\multicolumn{4}{l}{Note: $\star$ means that the constraint is
violated.}\\
\end{tabular} \label{tab4}
%\end{table*}
\end{sidewaystable}

\begin{sidewaystable}
%\begin{table*}[t]
  \centering
  \caption{Comparison of best result of different methods for solving the three bar truss problem}
  \begin{tabular} {llllll}
  \hline
                  & BAS algorithm                         &Cricket algorithm \cite{cbbec}  &\cite{edoua}  &\cite{appl2} &\cite{baana}\\
  \hline
$x_1$             & 0.788511192166172                     &0.788633                        &0.79500       &0.78867       &0.78863\\
$x_2$             & 0.408717503699073                    &0.408368                         &0.39500       &0.40902       &0.40838\\
$g_1(\mathbf{x})$ & -4.026245777222215e-06              &   -3.954291896146600e-07         &-0.00169      &-0.00029      & -3.057141794382545e-06\\
$g_2(\mathbf{x})$ &   -1.463570340396164                   &
-1.463965733302426           &-0.26124      &-0.26853 &
  -1.463953424351428                     \\
$g_3(\mathbf{x})$ &   -0.536433685849614                  & -0.536034662126764             &-0.74045      &-0.73176    & -0.536049632790367\\
$f(\mathbf{x})$   &    263.8963947787828                    &
263.8958968669962            &264.3000      &263.9716   &
263.8962483388589
\\
    \hline
\end{tabular} \label{tab5}
%\end{table*}
\end{sidewaystable}

\subsection{Speed reducer problem}
The speed reducer design optimization problem is described as
follows \cite{appl1}:
\begin{equation*}
\begin{aligned}
\text{min}&~f(\mathbf{x})=0.7854BH^2(3.3333Z^2+14.9334Z-43.0934)\\
&~~~~~~~~~~~~-1.508B(D^2_1+D^2_2)+
         7.4777(D^3_1+D^3_2)\\
         &~~~~~~~~~~~~+0.7854(L_1D^2_1+L_2D^2_2),\\
\text{subject to}~~& g_1(\mathbf{x})=\frac{27}{BH^2Z}-1\leq0,\\
&g_2(\mathbf{x})=\frac{397.5}{BH^2Z^2}-1\leq0,\\
&g_3(\mathbf{x})=\frac{1.93L^3_1}{HZD^4_1}-1\leq0,\\
&g_4(\mathbf{x})=\frac{1.93L^3_2}{HZD^4_2}-1\leq0,\\
&g_5(\mathbf{x})=\frac{1}{110D^3_1}\sqrt{(\frac{745L_1}{HZ})^2+16.9\times10^6}-1\leq0,\\
&g_6(\mathbf{x})=\frac{1}{85D^3_2}\sqrt{(\frac{745L_2}{HZ})^2+157.5\times10^6}-1\leq0,\\
&g_7(\mathbf{x})=\frac{HZ}{40}-1\leq0,\\
&g_8(\mathbf{x})=\frac{5H}{B}-1\leq0,\\
&g_9(\mathbf{x})=\frac{B}{12H}-1\leq0,\\
&g_{10}(\mathbf{x})=\frac{1.5D_1+1.9}{L_1}-1\leq0,\\
&g_{11}(\mathbf{x})=\frac{1.1D_2+1.9}{L_2}-1\leq0,\\
&2.6\leq B \leq 3.6,\\
&0.7\leq H \leq 0.8,\\
&17\leq Z \leq 28,\\
&7.3\leq L_1 \leq 8.3,\\
&7.8\leq L_2\leq 8.3,\\
&2.9\leq D_1\leq3.9,\\
&5.0\leq D_2\leq5.5,
\end{aligned}
\end{equation*}
where $B$ denotes the face width, $H$ denotes the module of the
teeth, $Z$ denotes the number of teeth on pinion, $L_1$ denotes the
length of the first shaft between bearings, $L_2$ denotes the length
of the second shaft between between bearings, $D_1$ denotes the
diameter of the first shaft, and $D_2$ denotes the the diameter  of
the second shaft.

We adopt the same approach as in the previous subsection to convert
the probloem into a form that can be addressed by the BAS algorithm.
With $\rho=10^6$, $\alpha = 0.8$, $d_0=0.001$, and $c=0.8$, and the
other settings being the same as in the previous subsection, the
best result obtained by the BAS algorithm among 100 runs with
$K_{max}=10,000$ is shown in Table \ref{tab4}. As seen from Table
\ref{tab4}, the solution given by the BAS algorithm can guarantee
the compliance with all the constraints with a optimal function
value being 3.012610927770214e+03. Although the other two algorithms
can generate better function values, some constraints are violated,
which means that the solutions are not feasible. From this point of
view, the BAS algorithm is better than the other two for solving
this problem.

\subsection{Three bar truss problem}
The three bar truss problem considered in this paper is described as
follows \cite{cbbec}:
\begin{equation*}
\begin{aligned}
\text{min}&~f(\mathbf{x})=100(2\sqrt{2}x_1+x_2),\\
\text{subject to}~~& g_1(\mathbf{x})=2\frac{\sqrt{2}x_1+x_2}{\sqrt{2}x^2_1+2x_1x_2}-2\leq0,\\
&g_2(\mathbf{x})=2\frac{x_2}{\sqrt{2}x^2_1+2x_1x_2}-2\leq0,\\
&g_3(\mathbf{x})=2\frac{1}{x_1+\sqrt{2}x_2}-2\leq0,
\end{aligned}
\end{equation*}
where $0<x_1<1$ and $0<x_2<1$. We employ the BAS algorithm to solve
the problem with         $\alpha = 0.8$, $d_0=0.01$, and $c=0.8$.
The comparison of the obtained best result with existing ones is
shown in Table \ref{tab5}. As seen from the table, the best result
obtained  by the BAS algortihm is very close to the those obtained
by the state-of-the-art, and all the constraints are satisfied.

\section{Conclusions}\label{sec.5}
In this paper, theoretical guarantee for the BAS algortihm has been
provided via the concept of convergence with probability 1. We have
also provided a quantitive analysis on the performance of the BAS
algorithm for finding global optima of seven representative test
functions based the measure called successful rate. The BAS
algorithm has been applied to solve three problems arising from
engineering applications, and the results have shown that the BAS
algorithm has a good performance.


\begin{thebibliography}{99}



\bibitem{basba}
X. Jiang and S. Li, BAS: Beetle antennae search algorithm for
optimization problems, Int. J. Robot. Control 1(1) (2018) 1--5.

\bibitem{anbas}
Z. Zhu, Z. Zhang, W. Man, X. Tong, J. Qiu, and F. Li, A new beetle
antennae search algorithm for multiobjective energy management in
microgrid, 13th IEEE Conf. Ind. Electron. Appl. (2018) 1599--1603.

\bibitem{asfcm}
X. Yin and Y. Ma, Aggregation service function chain mapping plan
based on beetle antennae search algorithm,  2nd Int. Conf. Telecom.
Comm. Eng. (2018) 225--230.

\bibitem{rosee}
C. Wang, C. Ren, B. Li, Y. Wang, and K. Wang, Research on
straightness error evaluation method based on search algorithm of
beetle, Int. Workshop  Adv. Manuf. Autom. (2018) 368--374.

\bibitem{onnub}
Y. Sun, J. Zhang, G. Li, Y. Wang, J. Sun, and C. Jiang, Optimized
neural network using beetle antennae search for predicting the
unconfined compressive strength of jet grouting coalcretes, Int. J.
Numer. Anal. Methods Geomech. 43(4) (2019) 801--813.

\bibitem{darod}
X. Lin, Y. Liu, and Y. Wang, Design and research of DC motor speed
control system based on improved BAS, Chinese Autom. Cong. (2018)
3701--3705.

\bibitem{doymo}
Y. Sun, J. Zhang, G. Li, G. Ma, Y. Huang, J. Sun, Y. Wang, B. Nener,
Determination of Young's modulus of jet grouted coalcretes using an
intelligent model, Eng. Geol. 252 (2019) 43--53.

\bibitem{popau}
J. Sun, J. Zhang, Y. Gu, Y. Huang, Y. Sun, and G. Ma, Prediction of
permeability and unconfined compressive strength of pervious
concrete using evolved support vector regression, Constr. Build.
Mater. 207 (2019) 440--449.











\bibitem{bsofs}
T. Chen, Y. Zhu, and J. Teng, Beetle swarm optimisation for solving
investment portfolio problems, J. Eng. 2018 (2018) 1600--1605.

\bibitem{aopso}
D. Song, Application of particle swarm optimization based on beetle
antennae search strategy in wireless sensor network coverage dianna
song, Advan. Intell. Syst. Res. 147 (2018) 1051--1054.

\bibitem{ahomo}
M. Lin and Q. Li, A Hybrid optimization method of beetle antennae
search algorithm and particle swarm optimization, Int. Conf. Elect.
Control Autom. Robot. (2018) 396--401.

\bibitem{mbrst}
F. J. Solis and R. J. B. Wets, Minimization by random search
technique, Math. Oper. Res. 6(1) (1981) 19--30.

\bibitem{def}
A. H. Gandomi and X. Yang, Chaotic bat algorithm, J. Comput. Sci. 5
(2014) 224--232.

\bibitem{yang1}
X. S. Yang, Test problems in optimization, in: Engineering
Optimization: An Introduction with Metaheuristic Applications, John
Wiley $\&$ Sons, (2010).

\bibitem{yang2}
M. Jamil and X. Yang, A literature survey of benchmark functions for
global optimization problems, Int. J. Math. Model.  Numeri. Opt.,
4(2) (2013) 150--194.

\bibitem{appl1}
X. S. Yang, M. Karamanoglu, S. Fong, Bat algorithm for topology
optimization in microelectronic applications, 1st Int. Conf. Future
Gen. Comm. Technol. IEEE (2012) 150--155.

\bibitem{appl2}
A. H. Gandomi, X. S. Yang, A. H. Alavi, Cuckoo search algorithm: a
metaheuristic approach to solve structural optimization problems,
Eng. Comput. 27, article DOI 10.1007/s00366-011-0241-y, (2011).


\bibitem{apl3}
M. Lin, J. Tsai, N. Hu, S. Chang, Design optimization of a speed
reducer using deterministic techniques, Math. Prob. Eng. 2013,
article ID 419043, (2013).

\bibitem{cbbec}
M. Canayaz, A. Karci, Cricket behaviour-based evolutionary
computation technique in solving engineering optimization problems,
Appl. Intell. 44 (2016) 362--376.

\bibitem{edoua}
 T. Ray, P. Saini, Engineering design optimization using a
swarm with an intelligent information sharing among individuals.
Eng. Optim. 33 (2001) 735--748.

\bibitem{baana}
X. S. Yang, A. H. Gandomi, Bat algorithm: a novel approach for
global engineering optimization, Eng. Comput. 29 (2012) 464--483.

\end{thebibliography}
\end{document}